\documentclass[runningheads]{llncs}

\usepackage[table,dvipsnames]{xcolor}
\usepackage[T1]{fontenc}
\usepackage{comment}
\usepackage{graphicx}
\usepackage{booktabs}
\usepackage{tikzducks}
\usepackage{amsmath}
\usepackage{hyperref}
\usepackage{tabularx}
\usepackage[ruled]{algorithm2e}
\usepackage{caption}
\usepackage{subcaption}
\usepackage{adjustbox}
\usepackage{cite}
\usepackage{soul}
\usepackage{orcidlink}
\usepackage{
tikz,
relsize,
tikz,
wrapfig
}
\usepackage{lipsum}

\hypersetup{
  colorlinks   = true, 
  urlcolor     = blue,
  linkcolor    = blue, 
  citecolor   = red 
}

\begin{document}

\title{\textit{FedRewind}: Rewinding Continual Model Exchange for Decentralized Federated Learning}

\titlerunning{FedRewind}

\author{Luca Palazzo$^{*\dagger}    $\inst{1}\orcidlink{0009-0008-2241-2643} \and
Matteo Pennisi$^*$\inst{1}\orcidlink{0000-0002-6721-4383}\and
Federica Proietto Salanitri\inst{1}\orcidlink{0000-0002-6122-4249} \and 
Giovanni Bellitto\inst{1}\orcidlink{0000-0002-1333-8348} \and
Simone Palazzo\inst{1}\orcidlink{0000-0002-2441-0982} \and
Concetto Spampinato\inst{1}\orcidlink{0000-0001-6653-2577}
}

\let\oldthefootnote=\thefootnote%
\def\thefootnote{*}\footnotetext{Equal contribution}
\def\thefootnote{$\dagger$}\footnotetext{Corresponding Author}
\let\thefootnote=\oldthefootnote%

\authorrunning{L. Palazzo et al.}

\institute{PeRCeiVe Lab, University of Catania, Italy. \\ \url{http://www.perceivelab.com/}}

\maketitle            

\begin{abstract}

In this paper, we present \textit{FedRewind}, a novel approach to decentralized federated learning that leverages model exchange among nodes to address the issue of data distribution shift. Drawing inspiration from continual learning (CL) principles and cognitive neuroscience theories for memory retention, \emph{FedRewind} implements a decentralized routing mechanism where nodes send/receive models to/from other nodes in the federation  to address spatial distribution challenges inherent in distributed learning (FL).  During local training, federation nodes periodically send their models back (i.e., \textit{rewind}) to the nodes they received them from for a limited number of iterations. This strategy reduces the distribution shift between nodes' data, leading to enhanced learning and generalization performance. 
We evaluate our method on multiple benchmarks, demonstrating its superiority over standard decentralized federated learning methods and those enforcing specific routing schemes within the federation. 
Furthermore, the combination of federated and continual learning concepts enables our method to tackle the more challenging federated continual learning task, with data shifts over both space and time, surpassing existing baselines.

\keywords{Decentralized Learning \and Continual Learning \and Federated Learning.}
\end{abstract}

\section{Introduction}
\label{sec:intro}
The proliferation of data across multiple distributed devices and locations has sparked significant interest in federated learning (FL), a paradigm that enables collaborative model training without the need to centralize data. Federated learning offers numerous benefits, including enhanced privacy and reduced communication costs. However, a fundamental challenge in FL is the non-i.i.d. (independent and identically distributed) nature of data across different nodes, which can lead to performance degradation due to data distribution shifts. This problem becomes even more pronounced in decentralized federated learning, where there is no central server to coordinate and aggregate updates, making the system less robust to heterogeneous data distributions.

Existing solutions in federated learning primarily focus on mitigating the effects of non-i.i.d. distributions through various aggregation and optimization techniques. Centralized federated learning approaches often rely on a central server to aggregate updates from all nodes, thereby smoothing out the differences in local data distributions~\cite{mcmahan2017communication, li2020federated, shoham2019overcoming}. Decentralized methods, instead, employ peer-to-peer communication and model averaging strategies to achieve consensus without a central entity~\cite{chang2018distributed, wink2021approach,beltran2023decentralized}. While these methods have shown promise, they often fall short in addressing the dynamic nature of data distribution shifts, particularly in environments featured by strong data imbalance~\cite{yang2024federated,zhu2022diurnal}.

Continual learning (CL)~\cite{mccloskey1989catastrophic,de2019continual,parisi2019continual}, a field that addresses the problem of learning from a stream of data that changes over time, offers valuable insights for handling distribution shifts with strong imbalance. CL methods are designed to prevent \emph{catastrophic forgetting}, which occurs when a model forgets previously learned information upon encountering new data, by maintaining knowledge across sequential learning tasks thourgh  either exposing the model to limited past experience~\cite{robins1995catastrophic,rebuffi2017icarl,buzzega2020dark, bellitto2022effects} or regularizing model parameters~\cite{kirkpatrick2017overcoming, zenke2017continual, boschini2022transfer} using previous knowledge, while learning new tasks. Although CL and FL address similar challenges, they operate in different contexts: CL deals with non-i.i.d. data over time, while FL addresses non-i.i.d. data across distributed nodes.

We here propose \emph{FedRewind}, a novel approach that integrates continual learning concepts into federated learning to address the limitations of existing FL methods. Our method involves decentralized nodes periodically exchanging their models and sending them back to the originating nodes for a limited number of iterations during local training. This exchange mechanism, inspired by continual learning strategies, aims to prevent overfitting on local data and enhance memory retention by periodically re-exposing models to previously seen data.

\textit{FedRewind}'s strategy also aligns with the cognitive neuroscience principle of \textit{testing effect}, which emphasizes the role of  active recall and retrieval practice for the enhancement of long-term memory.
The testing effect, in particular, demonstrates that memory retrieval processes (similar to our rewind strategy) significantly improve knowledge retention compared to simple re-exposure to information \cite{roediger2011critical, karpicke2011retrieval}. This phenomenon is underpinned by mechanisms such as elaborative retrieval and spreading activation, where active recall strengthens memory traces and facilitates the integration of new information into existing cognitive frameworks. 

By adapting cognitive neuroscience principles and continual learning concepts to the spatial distribution challenges of FL, \textit{FedRewind} aims to reduce the distribution shift between nodes, thus enhancing model performance and robustness.
We validate our claims on multiple benchmark datasets, demonstrating how \emph{FedRewind} leads to performance improvement over standard decentralized federated learning methods, as well as those that impose specific routing schemes within the federation. Furthermore, the combination of federated and continual learning concepts enables our method to effectively address the federated continual learning problem, where data shifts occur over both space and time, outperforming existing baselines. 
Our results, cumulatively, indicate that this decentralized and iterative model exchange approach offers a robust solution to the challenges posed by non-i.i.d. data in federated learning environments.

\section{Related Work}

\textbf{Federated learning (FL)} has emerged as a new paradigm within distributed machine learning, addressing the challenge of data privacy. Drawing upon the foundational work of McMahan et al.~\cite{mcmahan2017communication}, FL facilitates collaborative model training while ensuring node data remains secure on their local devices.

A typical FL setting features a central server that orchestrates the learning process. This server distributes a global model to a pool of participating nodes, which use their private data for local updates on the received model. Subsequently, the nodes transmit their local updates back to the central server, which aggregates them to refine the global model. This iterative process of distributing, updating, and aggregating the model persists until a satisfactory level of convergence is achieved.The most common aggregation technique is FedAvg~\cite{mcmahan2017communication}, that simply averages the local model parameters received from all nodes. More sophisticated aggregation methods have been proposed by adding a regularization term\cite{li2020federated} or leveraging knowledge distillation\cite{zhu2021data}. 
Another branch of FL, namely Personalized Federated Learning, has the primary objective to improve the performances w.r.t. only the single node distribution. FedBN\cite{li2021fedbn} achieves this goal by preserving the batch-norm layers of each node while FedProto\cite{tan2022fedproto} aggregates only the prototypes while the models are kept on each node.

While a central server simplifies the communication protocols, especially for large-scale deployments, its presence introduces specific limitations. Firstly, it creates a single point of failure, posing a vulnerability to system robustness if the server becomes unavailable. Secondly, as the number of participating nodes increases, the central server can become a bottleneck, hindering communication efficiency~\cite{lian2017can}. Finally, the very presence of a central server that aggregates data might not be desirable or even feasible in certain collaborative learning scenarios. This is particularly true for scenarios that prioritize robust privacy guarantees or involve geographically dispersed participants with limited or unreliable network connectivity.

This work investigates also decentralized federated learning, which, conversely to centralized approaches, relies on peer-to-peer communication between nodes\cite{chang2018distributed,kalra2023decentralized}. This approach eliminates the single point of failure and enhances privacy guarantees, but introduces additional complexity in terms of communication protocols and achieving convergence among local models on all devices.\\

Since \textit{FedRewind} leverages concepts from \textbf{continual learning (CL)}, we provide a brief overview of existing methods related to the strategies we employ to retain knowledge from past learning rounds.
Continual learning~\cite{de2019continual, parisi2019continual} is a field of machine learning that seeks to bridge the gap between the incremental learning observed in humans and the limitations of neural networks. McCloskey and Cohen~\cite{mccloskey1989catastrophic} identified the phenomenon of ``catastrophic forgetting'', where neural networks lose previously acquired knowledge upon encountering substantial shifts in the input distribution.

To address catastrophic forgetting, various mitigation strategies have been proposed. These include the introduction of appropriate regularization terms~\cite{kirkpatrick2017overcoming, zenke2017continual}, the development of specialized network architectures~\cite{schwarz2018progress, mallya2018packnet}, and the use of rehearsal mechanisms that leverage a limited set of previously encountered data points~\cite{robins1995catastrophic, rebuffi2017icarl, buzzega2020dark}.

\textit{FedRewind} adopts a hybrid approach, combining elements of both regularization and rehearsal strategies. Unlike traditional methods, it does not use any buffer. Instead, during training rounds on local nodes, the model is periodically sent back to previous nodes for regularization. This process helps address data shifts across nodes, thereby mitigating potential forgetting.\\

\textbf{Federated continual learning (FCL)} combines the paradigms of federated learning (FL) and continual learning (CL), enabling it to address the challenge of distributed data that, at the same time, undergo continual change over time~\cite{dong2022federated, yao2020continual, yoon2021federated}. 
However, initial efforts in this area compromised data privacy by requiring the storage of training samples on the central server~\cite{yao2020continual}. Recent advancements in FCL prioritize data privacy by advocating for the storage of only perturbed images for replay purposes~\cite{dong2022federated}. FedWEIT~\cite{yoon2021federated}, instead, tackled the problem by decomposing network weights but necessitates data replay buffers.
Recently, FedSpace~\cite{shenaj2023asynchronous} has been developed to overcome the limitations of current methods. It leverages class prototypes within the feature space and employs contrastive learning to preserve prior knowledge and reduce divergence between the behaviors of different federation nodes.

Our \emph{FedRewind} strategy is complementary to approaches like FedSpace, enhancing their capabilities by mitigating typical overfitting in distributed learning, particularly in cases of high class imbalance and strongly non-iid data distribution among nodes. Specifically, \emph{FedRewind} addresses these issues by transferring models between nodes to enforce iid-ness on data, rather than relying on data storage. This method significantly reduces privacy concerns associated with storing sensitive training data. By periodically sending the model back to previous nodes, we maintain knowledge across sequential tasks and enforce regularization, thus reducing node overfitting, while enhancing both privacy (no need for data sharing) and scalability.

\section{Method}
\label{sec:method}
\begin{figure}[t]
  \centering
  \includegraphics[width=0.98\textwidth]{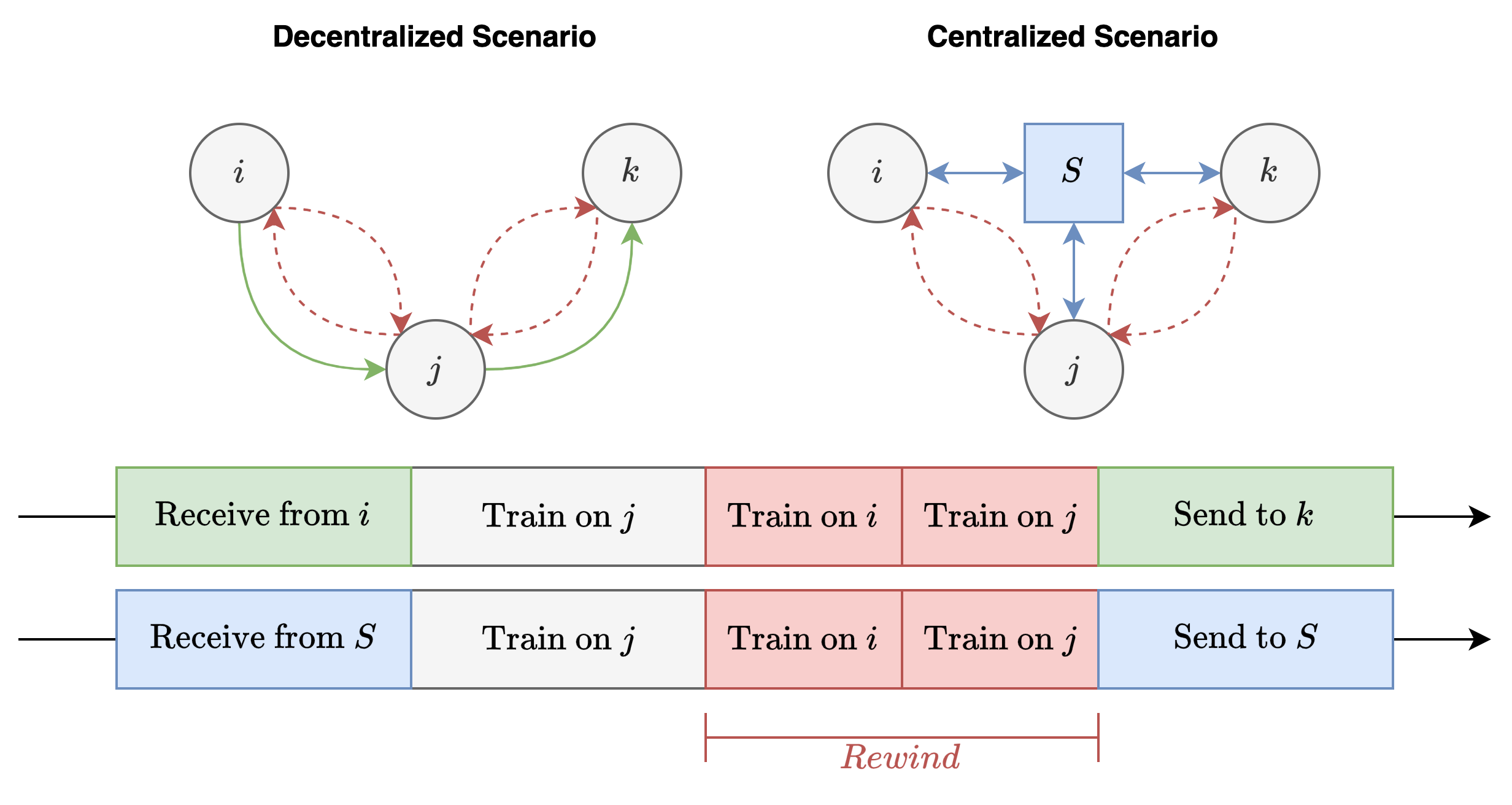}
  \caption{\textbf{Rewind strategy.} The model received and trained on the current node is sent back to its source node for a brief fine-tuning. The model then returns to the node and continue its training before the start of a new federated round.}
  \label{fig:method}
\end{figure}

In federated learning, a collection of nodes collaboratively train a shared model while keeping their data localized. This approach ensures data privacy but introduces challenges related to effective knowledge sharing across distributed and sequential learning tasks. To address these challenges, we propose a novel ``rewind'' strategy. This section introduces the approach and key concepts, describes the rewind method in detail, and provides the corresponding pseudo-code.

\paragraph{{Federated Learning.}} Federated learning is a collaborative machine learning approach where multiple nodes ($N$ nodes) train a shared model without centralizing their data. Each node updates its model using local data and shares the model updates rather than the data itself, ensuring data privacy.

\paragraph{{Decentralized Federated Learning.}} In this framework, nodes communicate directly with each other without a central server. We consider two modes of communication:
\begin{itemize}
    \item \textbf{Random Communication (RWT)}: Nodes select a random source node (for the incoming model) and a random destination node (for the outcoming model) for information exchange in each round.
    \item \textbf{Cyclic Communication (CWT)}: Each node communicates with the same predetermined source and destination nodes in every round, as described in \cite{chang2018distributed}.
\end{itemize}

\paragraph{{Centralized Federated Learning.}} In this framework, a central server coordinates the training process. Nodes send their local model updates to the server, which aggregates them to form a global model.

\paragraph{{Training Rounds.}} Defined as a block of training where all nodes complete training for \( E \) epochs. At the end of each round, model exchange across all federated nodes is carried out.

\subsection{The Rewind Strategy}

To improve knowledge sharing in federated learning, we introduce the ``rewind'' strategy, described in Fig.~\ref{fig:method}. This method involves temporarily reverting the model to a previous node to rehearse prior knowledge, thus preserving data privacy. We apply this strategy on both centralized and decentralized federated learning. 

\paragraph{Decentralized federated learning.} During each communication round, a generic node \( C_j \) receives a model \( M_i \), parameterized by $\theta_i$, from a source node \( C_i \), trains it on its local dataset \( D_j \), and forwards it to another node \( C_k \). The standard training process on node $C_j$ for model \( M_j \) parameterized by $\theta_j$ at round \( t \) on dataset $D_j$ is given by:

\begin{equation}
M_j^{(t)} = \text{Train}_{D_j, E}(M_i^{(t-1)}) = \theta_i^{(t-1)} - \eta \sum_{0}^{E-1} \nabla L(\theta_i^{(t-1)}, D_j)
\end{equation}

where \( E \) denotes the number of epochs for a single federation round, and \( L \) is a generic loss function.

To enhance knowledge retention, we introduce a fractional computation budget parameterized by \(\lambda\) for retraining the model on its origin node before continuing training on the current node. This modifies the training equation as follows:

\begin{equation}
M_j^{(t)} = \text{Train}_{D_j, \lambda \cdot E} \left( \text{Train}_{D_i, \lambda \cdot E} \left( \text{Train}_{D_j, (1- 2\lambda) \cdot E} \left( M_i^{(t-1)} \right) \right) \right)
\end{equation}

where $\lambda$ denote the fraction of the budget allocated for rewinding.

\paragraph{Centralized Federated Learning.} In this scenario, a central server \( S \) aggregates models received from nodes at each communication round. The model computed by the server \( M_s^{(t)} \) at round \( t \) is defined as:

\begin{equation}
M_s^{(t)} = agg(\{M_j^{(t)} \mid j \in \{1, 2, \ldots, N\}\})
\end{equation}

where \( agg(\cdot) \) represents a generic aggregation function employed by the server. Applying the rewind strategy, the training process for a generic node \( C_j \) is modified to:

\begin{equation}
M_j^{(t)} = \text{Train}_{D_j, \lambda \cdot E} \left( \text{Train}_{D_i, \lambda \cdot E} \left( \text{Train}_{D_j, (1- 2\lambda) \cdot E} \left( M_s^{(t-1)} \right) \right) \right)
\end{equation}

By leveraging inter-node communication and the rewind strategy, federated learning—whether decentralized or centralized—can more effectively retain knowledge across different tasks. This approach ensures that the shared model benefits from the distributed data while maintaining privacy and improving performance.

The pseudo-code of the rewind strategy is reported in Alg.~\ref{alg:rewind_decentr} and Alg.~\ref{alg:rewind_centr} respectively for the decentralized and centralized scenario.

\begin{algorithm}[H]

\definecolor{gray}{rgb}{0.4, 0.4, 0.4}

\SetCommentSty{mycommfont}
\newcommand\mycommfont[1]{\footnotesize\ttfamily\textcolor{gray}{#1}}
\caption{Decentralized Federated Learning with Rewind Strategy.}
\label{alg:rewind_decentr}
\SetAlgoNlRelativeSize{-1}
\SetAlgoNlRelativeSize{0}
\SetKwComment{Comment}{/* }{ */}

\BlankLine

\KwIn{$N$ nodes, initial model $M_0$, epochs $E$, fractional budget $\lambda$}

\For{each round $t$ in $1$ to $T$}{
    \For{each node $C_j$ in $N$}{
        \tcp{Receive model from source node $C_i$ }
        $M_i^{(t-1)} \gets$ receive\_model($C_i$)\;
        \tcp{Train on current node's dataset $D_j$}
        $M_\text{intermediate} \gets$ Train($M_\text{rewind}$, $D_j$, $(1 - 2\lambda) \cdot E$)\;
        \tcp{Rewind phase: Train on source node's dataset $D_i$}
        $M_\text{rewind} \gets$ Train($M_i^{(t-1)}$, $D_i$, $\lambda \cdot E$)\;
        \tcp{Finish training on current node's dataset $D_j$}
        $M_j^{(t)} \gets$ Train($M_\text{intermediate}$, $D_j$, $\lambda \cdot E$)\;
        \tcp{Send model to destination node $C_k$}
        send\_model($C_j$, $M_j^{(t)}$)\;
    }
}
\end{algorithm}

\begin{algorithm}[H]
\definecolor{gray}{rgb}{0.4, 0.4, 0.4}
\SetCommentSty{mycommfont}
\newcommand\mycommfont[1]{\footnotesize\ttfamily\textcolor{gray}{#1}}
\caption{Centralized Federated Learning with Rewind Strategy.}
\label{alg:rewind_centr}
\SetAlgoNlRelativeSize{-1}
\SetAlgoNlRelativeSize{0}
\SetKwComment{Comment}{/* }{ */}
\KwIn{$N$ nodes, initial model $M_0$, epochs $E$, fractional budget $\lambda$}
\BlankLine
\For{each round $t$ in $1$ to $T$}{
    \For{each node $C_j$ in $N$}{
        \tcp{Receive aggregated model from the server}
        $M_s^{(t-1)} \gets$ receive\_model(server)\;
        \tcp{Train on current node's dataset $D_j$}
        $M_\text{intermediate} \gets$ Train($M_\text{rewind}$, $D_j$, $(1 - 2\lambda) \cdot E$)\;
        \tcp{Rewind phase: Train on previous node's dataset $D_i$}
        $M_\text{rewind} \gets$ Train($M_s^{(t-1)}$, $D_i$, $\lambda \cdot E$)\;
        \tcp{Finish training on current node's dataset $D_j$}
        $M_j^{(t)} \gets$ Train($M_\text{intermediate}$, $D_j$, $\lambda \cdot E$)\;
        \tcp{Send model to the server}
        send\_model(server, $M_j^{(t)}$)\;
    }
    \tcp{Server aggregates models from all nodes}
    $M_s^{(t)} \gets$ aggregate\_models(\{$M_j^{(t)} \mid j \in 1 \text{ to } N$\})\;
}

\end{algorithm}

\section{Results}
\label{sec:results}
\subsection{Federated Learning Performance}

\noindent \textbf{Experimental settings.} To evaluate the effectiveness of \textit{FedRewind}, we simulate different federated learning scenarios using three benchmarks (generally employed for testing FL methods), namely MNIST~\cite{deng2012mnist}, CIFAR10~\cite{krizhevsky2009learning} and CIFAR100~\cite{krizhevsky2009learning}.
Data is distributed across nodes according to a non-independent and identically distributed (non-IID) scheme. This distribution is achieved by applying a Dirichlet distribution, as in previous work~\cite{li2021model,wang2020federated,yurochkin2019bayesian}, parameterized by $\alpha_{dir}$, which serves as a measure of the degree of data heterogeneity across nodes; a lower $\alpha_{dir}$ value indicates a more pronounced imbalance in data distribution across nodes. \\
Our experimental settings include 50 communication rounds and two configurations based on the number of nodes in the federation: one with 10 nodes and another with 50 nodes.
During each round, in each node, we perform local training for 10 epochs (E). For the rewind experiments, we set the rewind hyperparameter $\lambda = 0.1$. This configuration determines a training procedure where, within the given E=10 epochs, 8 epochs are dedicated to the local training on the current node's data, followed by one epoch on the previous node's data, and concluding with a final epoch on the current node's data.
All experiments are carried out using the ResNet18 architecture~\cite{he2016deep}, pre-trained on ImageNet~\cite{5206848}, optimized using Stochastic Gradient Descent (SGD) with a learning rate of 0.001. \\

\noindent \textbf{Metrics.}
 In decentralized federated learning (FL), each node creates a distinct local model, unlike in the centralized FL paradigm, which results in a single global model at the end of the training phase. 
 To quantify the aggregated performance and generalization capability of the federation, we propose the \emph{Federation Accuracy (FA)} metric. This metric is calculated by testing all the node models within the federation against all the private test sets and then computing the mean accuracy. Given a federation of size \( N \), our metric is defined as follows:

\begin{equation}
\text{FA} := \frac{1}{N \times N} \sum_{i=1}^{N} \sum_{j=1}^{N} \text{Acc}(M_i, D_j^{test})
\end{equation}

where $\text{Acc}(M_i, S_j)$ is the accuracy of model $M_i$ on the private test set $D_j^{test}$ of node $j$.
Similarly we define the \emph{Federation Fairness (FF)} that measures how much the performance of the nodes changes across the federation (e.g. the standard deviation of the accuracy of nodes):

\begin{equation}
\text{FF} := \sqrt{ \frac{1}{(N \times N) - 1} \sum_{i=1}^{N} \sum_{j=1}^{N} (\text{Acc}(M_i, D_j^{test}) - \text{FA})^2}
\end{equation}

Moreover, our objective is not only to improve the overall generalization across the federation, but also to enhance performance of each individual node on its private dataset. To measure this performance, we define the \emph{Personalized Federation Accuracy (PFA)} metric as:

\begin{equation}
\text{PFA} := \frac{1}{N} \sum_{i=1}^{N} \text{Acc}(M_i, D_i^{test}).
\end{equation}

These three metrics, PA, FF and PFA, allow us to capture both the generalization capabilities of the entire federation and the performance improvements from the perspective of individual nodes.\\

\textbf{Baselines.} We test our approach in combination to existing FL strategies, applying it to CWT~\cite{chang2018distributed}, RWT, and FedAvg~\cite{mcmahan2017communication}. CWT employs a static cyclic model transfer between rounds, while RWT is our modified version of CWT, featuring random communication between nodes in each round. 

We also assess our approach in two reference scenarios: the \emph{Joint} and \emph{Standalone} settings. The \emph{Joint} setting represents an optimal condition where all data from the federation is consolidated and utilized for training on a single node, thereby establishing an upper bound on performance.

In contrast, the \emph{Standalone} setting assumes that each node trains its model independently, with no communication or data sharing between nodes. This setting generally sets a lower bound on performance, particularly when the data distribution between nodes is highly non-IID.\\

\noindent \textbf{Results.} We begin our evaluation by testing FedRewind performance on the two scenarios: with 10 nodes and with 50 nodes, both under a strongly non-IID scenario with $\alpha_{dir}=0.25$. Table \ref{tab:performance} presents the results in terms of Federation Accuracy (FA) across the three benchmarks, showing that the \emph{rewind} strategy consistently achieves higher accuracy. This improvement is especially pronounced in the 50-node scenario, which poses greater complexity and challenge due to its larger and more heterogeneous node distribution. Similarly, Table~\ref{tab:fairness} presents the results in terms of Federation Fairness (FF), demonstrating that the implementation of the rewind strategy consistently reduces the standard deviation of the accuracy of nodes, thus enhancing the generalization capabilities of the federation.

The Personalized Federation Accuracy (PFA) results, shown in Table \ref{tab:performance_pers}, further demonstrate the benefits of our rewind strategy at the node level. The strategy’s effectiveness is evident, as it consistently enhances PFA across various datasets and federation scales.
We speculate that the rewind mechanism acts as an effective regularizer, mitigating overfitting to a node’s local dataset. In a non-IID setting, where the tendency to overfit to local data patterns is high, this regularization effect is crucial. By periodically rewinding and retraining with data from other nodes, the models are exposed to a more diverse data distribution, promoting more generalized representation learning. It is also noteworthy that the impact of our rewind strategy on personalization performance for FedAvg is relatively lower compared to CWT and RWT. This might be due to the aggregation step in FedAvg, which tends to smooth out the specificities of local models trained on non-IID data.

The results from the standalone setting highlight the limitations of training models in isolation, especially under non-IID conditions. As the number of nodes increases, standalone models generally perform poorly, struggling to learn representative features without the diversity of data from other nodes. This aligns with our expectations, as the standalone setting misses the collaborative benefits of federated learning.

\begin{table}[ht]
\centering
\setlength{\tabcolsep}{8pt}
\rowcolors{2}{gray!10}{white}
\footnotesize{
\begin{tabular}{l|ccc|ccc}
\toprule
\textbf{Method}         & \multicolumn{3}{c}{\textbf{10 Nodes}}                 & \multicolumn{3}{c}{\textbf{50 Nodes}}                 \\ \midrule
        & \textbf{MNIST}    &  \textbf{C10}    &   \textbf{C100}   &   \textbf{MNIST}  &   \textbf{C10} & \textbf{C100} \\
                        \midrule
Joint       &   99.22           &   78.53               &   53.13           &   99.22           &   78.53               &   53.13               \\
Standalone  &   69.19           &   33.23               &   19.07           &   49.91           &   26.85   &
                11.08               \\ \midrule
FedAVG      &   95.43           &   51.19               &   39.89           &   87.47           &   53.26   &
                37.29           \\
\hspace{0.2 cm }\textbf{$\hookrightarrow$Rewind}    &   \textbf{97.59}  & \textbf{59.27}        & \textbf{41.06}  &   \textbf{91.77}    &
                \textbf{58.84}  & \textbf{39.90}        \\
\midrule
CWT         &   93.09           &   45.81               &   34.02           &    88.27          &   40.00   &
            24.90           \\
\hspace{0.2 cm }\textbf{$\hookrightarrow$Rewind}       &   \textbf{96.19}  &   \textbf{55.57}  &   \textbf{37.44}      &   \textbf{92.79}  &   \textbf{47.83} &     \textbf{27.51}              \\
\midrule

RWT         &   94.93           &   44.16               &   32.08               &    86.66      &   38.75                      &     24.42              \\
\hspace{0.2 cm }\textbf{$\hookrightarrow$Rewind}       &   \textbf{97.42}  &   \textbf{52.34}             &     \textbf{36.72}             &    \textbf{87.01}            &    \textbf{45.85}             &     \textbf{27.40}              \\

\bottomrule
\end{tabular}
}
\caption{\textbf{Federation Accuracy} in a non-IID setting ($\alpha_{dir} = 0.25$) for the MNIST, CIFAR-10, and CIFAR-100 benchmarks, organized across 10 and 50 nodes.}
\label{tab:performance}
\end{table}

\begin{table}[ht]
\centering
\setlength{\tabcolsep}{8pt}
\rowcolors{2}{gray!10}{white}
\footnotesize{
\begin{tabular}{l|ccc|ccc}
\toprule
\textbf{Method}         & \multicolumn{3}{c}{\textbf{10 Nodes}}                 & \multicolumn{3}{c}{\textbf{50 Nodes}}                 \\ \midrule
        & \textbf{MNIST}    &  \textbf{C10}    &   \textbf{C100}   &   \textbf{MNIST}  &   \textbf{C10} & \textbf{C100} \\
                        \midrule
Joint       &   N/A           &   N/A               &   N/A           &   N/A           &   N/A               &   N/A               \\
Standalone  &   27.02           &   25.47               &   12.15           &   28.97           &   20.59   &
                11.08               \\ \midrule
FedAVG      &   3.81           &   13.37               &   10.33           &   13.18           &   11.21   &
                5.33           \\
\hspace{0.2 cm }\textbf{$\hookrightarrow$Rewind}    &   \textbf{1.50}  & \textbf{12.75}        & \textbf{10.09}  &   \textbf{9.03}    &
                \textbf{10.10}  & \textbf{5.16}        \\
\midrule
CWT         &   8.52           &   24.21               &   11.22           &    14.18          &   21.36   &
            6.85           \\
\hspace{0.2 cm }\textbf{$\hookrightarrow$Rewind}       &   \textbf{4.78}  &   \textbf{19.53}  &   \textbf{10.07}      &   \textbf{8.48}  &   \textbf{18.84} &     \textbf{6.72}              \\
\midrule

RWT         &   4.27           &   24.95               &   10.84               &    17.54      &   21.26                      &     6.75              \\
\hspace{0.2 cm }\textbf{$\hookrightarrow$Rewind}       &   \textbf{1.82}  &   \textbf{21.98}             &     \textbf{10.06}             &    \textbf{14.68}            &    \textbf{19.01}             &     \textbf{6.69}              \\

\bottomrule
\end{tabular}
}
\caption{\textbf{Federation Fairness} in a non-IID setting ($\alpha_{dir} = 0.25$) for the MNIST, CIFAR-10, and CIFAR-100 benchmarks, organized across 10 and 50 nodes.}
\label{tab:fairness}
\end{table}

\begin{table}[ht]
\centering
\setlength{\tabcolsep}{8pt}
\rowcolors{2}{gray!10}{white}
\footnotesize{
\begin{tabular}{l|ccc|ccc}
\toprule
\textbf{Method}         & \multicolumn{3}{c}{\textbf{10 Nodes}}                 & \multicolumn{3}{c}{\textbf{50 Nodes}}                 \\ \midrule
        & \textbf{MNIST}    &  \textbf{C10}    &   \textbf{C100}   &   \textbf{MNIST}  &   \textbf{C10} & \textbf{C100} \\
                        \midrule
Joint       & 99.22           &   78.53               &   53.13  &     99.22           &   78.53               &   53.13\\
Standalone  &   98.69           &    83.96              &    53.49          &   96.20        &  77.65   &    41.73\\ 
\midrule
FedAVG      &   97.09           &    \textbf{56.15}              &    37.94          &    83.83          &  45.88    &  32.54\\
\hspace{0.2 cm }\textbf{$\hookrightarrow$Rewind}    &   \textbf{97.38}           &    54.39              &    \textbf{38.71}          &    \textbf{85.79}          &  \textbf{47.99}    &  \textbf{35.62} \\
\midrule
CWT         &   94.82           &    38.01              &    30.76          &    86.74          &   38.90   &  24.72  \\
\hspace{0.2 cm }\textbf{$\hookrightarrow$Rewind}       &   \textbf{95.65}           &    \textbf{48.85}              &    \textbf{34.34}          &    \textbf{93.40}          &   \textbf{46.29}   &  \textbf{26.58}  \\
\midrule

RWT         &   92.95           &    43.48              &    29.36          &   85.43           &   33.64   &  23.90\\
\hspace{0.2 cm }\textbf{$\hookrightarrow$Rewind}       &   \textbf{96.61}           &    \textbf{45.95}              &    \textbf{35.50}          &  \textbf{85.79}            &   \textbf{47.62}   &   \textbf{27.59}\\ 

\bottomrule
\end{tabular}
}
\caption{\textbf{Personalized Federation Accuracy} in a non-IID setting ($\alpha_{dir} = 0.25$) for the MNIST, CIFAR-10, and CIFAR-100 benchmarks, organized across 10 and 50 nodes.}
\label{tab:performance_pers}
\end{table}

\begin{figure}[htbp]
    \centering
    \includegraphics[width=\textwidth]{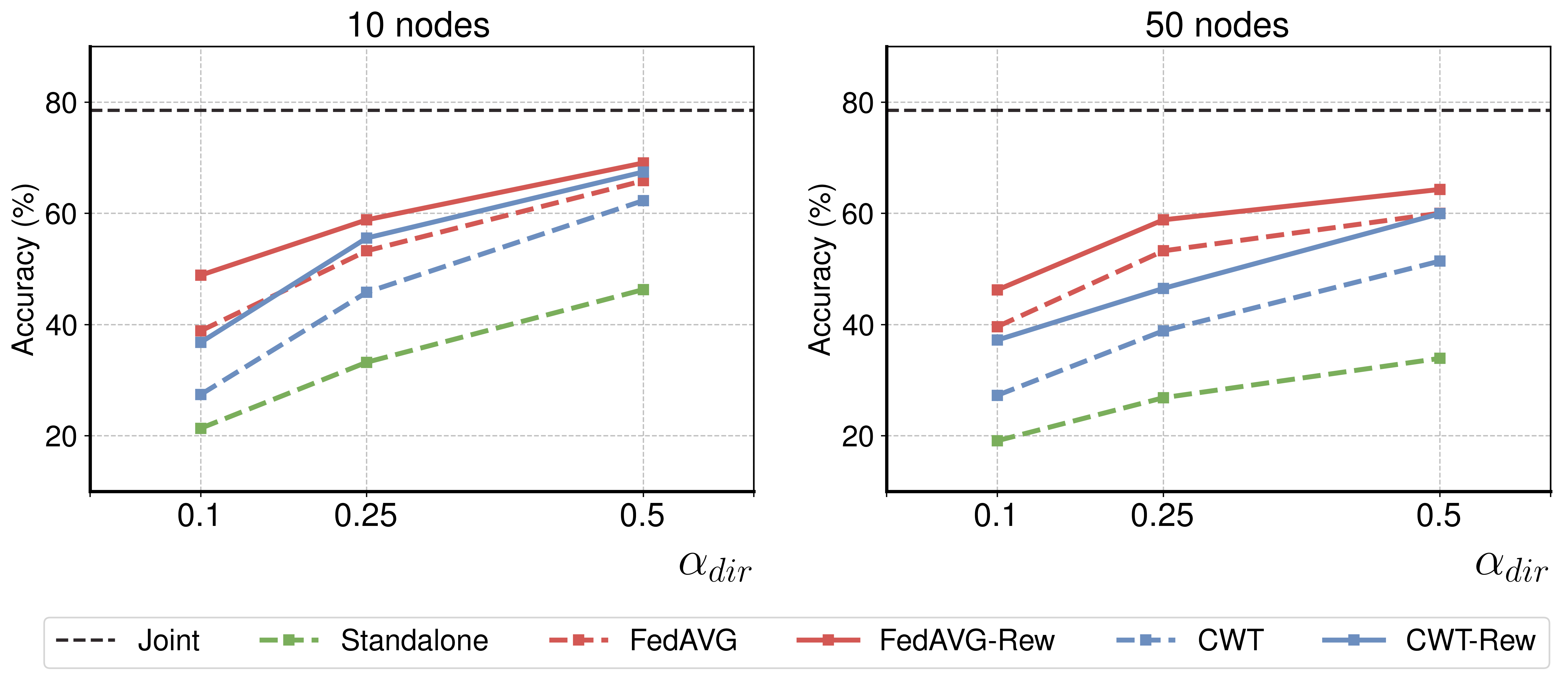}
    \caption{\textbf{Performance at different
degrees of data heterogeneity} ($\alpha_{dir}$) on CIFAR-10 for 10 (left) and 50 (right) nodes. }
    \label{fig:noniid}
\end{figure}

We further evaluated the robustness of our rewind strategy under varying degrees of data heterogeneity. Using the CIFAR-10 dataset, we measured federation accuracy with the Dirichlet coefficient ($\alpha_{dir}$) ranging from $0.1$, indicating extreme non-IID conditions, to $0.5$, representing a less skewed data distribution among nodes. The results, shown in Fig.~\ref{fig:noniid}, demonstrate that our rewind strategy consistently enhances FA, with the highest gain obtained at $\alpha_{dir}=0.1$, the most challenging setting. This suggests that the rewind strategy is particularly effective in environments with high data distribution skewness. As $\alpha_{dir}$ increases, \emph{FedRewind} continues to provide benefits, though they are less pronounced.
These findings collectively demonstrate that the rewind strategy is a robust method for federated learning, capable of enhancing model performance in diverse data conditions. Its consistent performance across different levels of non-IIDness, as shown in Fig.~\ref{fig:noniid}, suggests reliable applicability in real-world federated settings where data distributions vary widely. 

We finally verify whether the enhanced performance is due to rewinding to the previous node or to any other node. Thus, we compared our rewind strategy to a random rewinding one, where the model is sent to a random node of the federation.  
Table~\ref{tab:ablation} shows the performance of the two strategies when combined to CWT and RWT, highlighting how rewinding to the previous node in the communication chain is more effective than using a random node. It has to be noted that, in RWT, the performance increase is slightly lower because the preceding node changes at each communication round, slightly reducing the benefits of rewinding.

\begin{table}[ht]
\centering
\setlength{\tabcolsep}{8pt}
\rowcolors{2}{gray!10}{white}
\begin{tabular}{l|cccc}
\hline
\textbf{Method}                                        & \multicolumn{2}{c}{\textbf{CIFAR-10}}   & \multicolumn{2}{c}{\textbf{CIFAR-100}}    \\ \hline
                                                       & \textbf{10 Nodes} & \textbf{50 Nodes}   & \textbf{10 Nodes} & \textbf{50 Nodes} \\
\midrule
                       
CWT                                                    & 45.81             & 40.00               & 34.02             & 24.90  \\
\hspace{0.2 cm }\textbf{$\hookrightarrow$RandRewind}   & 51.59             & 45.76               & 36.58            & \textbf{27.60}  \\
\hspace{0.2 cm }\textbf{$\hookrightarrow$Rewind}       & \textbf{55.57}             & \textbf{47.83}               & \textbf{37.44}             & 27.51  \\
\midrule
RWT                                                    & 44.16             & 38.75               & 32.08             & 24.42  \\
\hspace{0.2 cm }\textbf{$\hookrightarrow$RandRewind}   & 50.83             & 45.62               & 36.62             & \textbf{27.40}  \\
\hspace{0.2 cm }\textbf{$\hookrightarrow$Rewind}       & \textbf{52.83}             & \textbf{45.85}               & \textbf{36.72}             & 27.09  \\
\bottomrule
\end{tabular}
\caption{\textbf{Comparison between different rewinding  strategies}. RandRewind sends the model back to a random done instead of the previous node as Rewind does.}
\label{tab:ablation}
\end{table}

\begin{figure}[t]
  \centering
  \begin{minipage}[c]{0.58\textwidth}
    \centering
    \includegraphics[width=\textwidth]{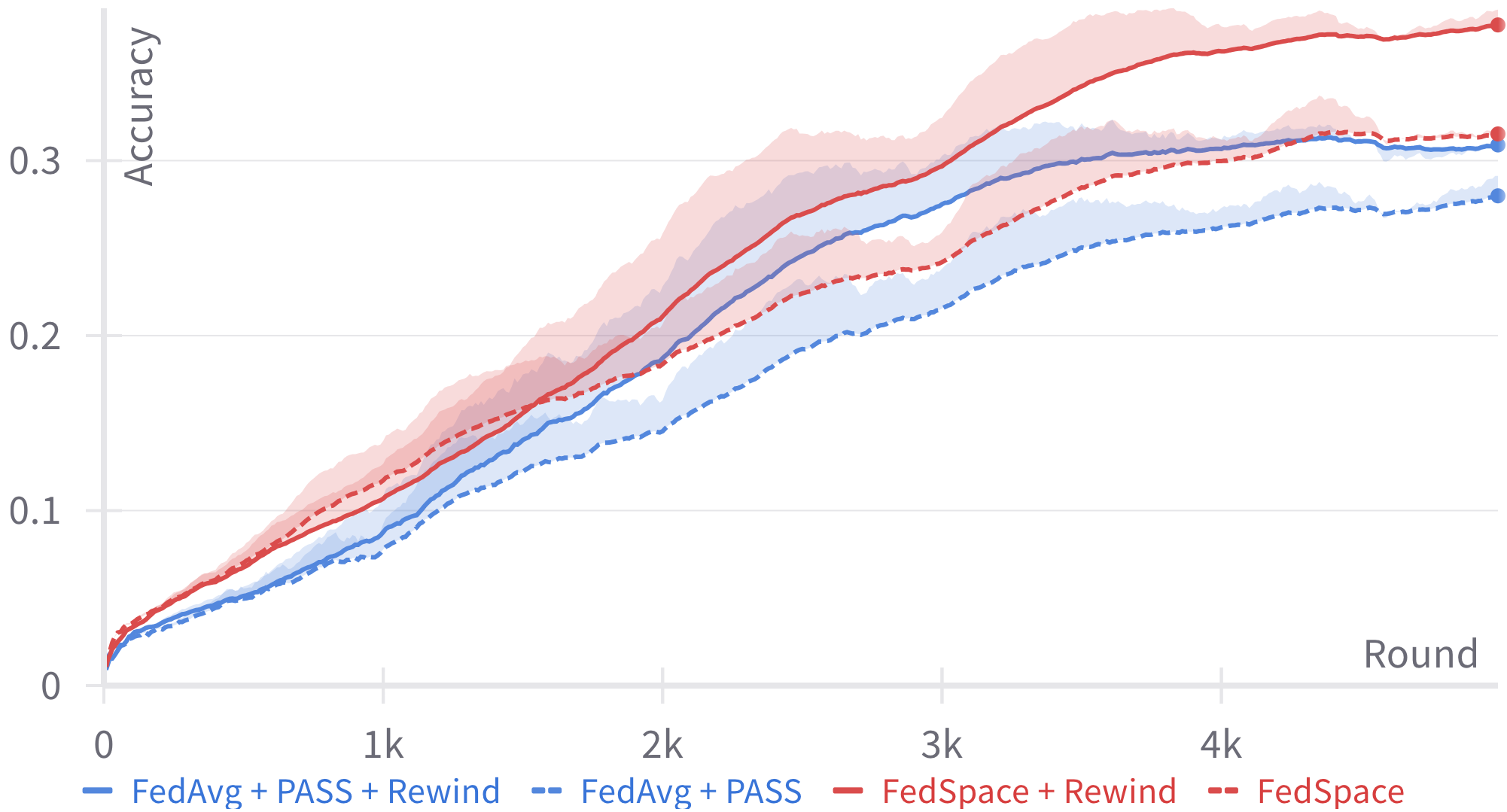}
    \captionof{figure}{\textbf{Training trend of rewind strategy in AFCL}}
    \label{fig:fedspace}
  \end{minipage}
  \hfill
  \begin{minipage}[c]{0.38\textwidth}
    \centering
    \footnotesize{
    \setlength{\tabcolsep}{2pt}
    \rowcolors{2}{gray!10}{white}
    \vspace{2.3em}
    \begin{tabular}{l|c}
    \toprule
         \textbf{Method} & \textbf{Accuracy}\\
         \midrule
         FedAvg + PASS & 29.70 \\
         \hspace{0.2cm}\textbf{$\hookrightarrow$Rewind} & \textbf{33.46}\\
         FedSpace & 35.53\\
         \hspace{0.2cm}\textbf{$\hookrightarrow$Rewind} & \textbf{39.40}\\
         \bottomrule
    \end{tabular}}
    \vspace{2.3em}
    \captionof{table}{\textbf{Results of rewind strategy in AFCL}}
    \label{tab:continual}
  \end{minipage}
\end{figure}

\subsection{Continual Federated Learning}

We also evaluate \emph{FedRewind} within the complex context of Asynchronous Federated Continual Learning (AFCL)~\cite{shenaj2023asynchronous}, where data is not only distributed across multiple nodes (as in federated learning) but also subject to changing distributions over time (as in continual learning). In this asynchronous setting, each node independently progresses through its continual learning tasks, creating unique distribution shifts at different times. 
We argue that the \emph{rewind} strategy is particularly advantageous in AFCL scenarios, as rewinding on another node can mitigate the exacerbated problem of forgetting.

To test this hypothesis, we implement our strategy on top of the current state-of-the-art approach for AFCL, FedSpace~\cite{shenaj2023asynchronous}. We replicate their experimental setup, using CIFAR100 divided into 10 tasks of 10 classes each, and maintain the same hyperparameters, except for the number of epochs per round, and without any pretraining. Specifically, we rerun their experiments with $E=3$ because the default value of 1 was incompatible with the rewind strategy. 
The experimental results and trends are detailed in Table~\ref{tab:continual} and Figure~\ref{fig:fedspace}, respectively, where the \emph{rewind} strategy is integrated into FedSpace. Additionally, we applied our proposed strategy to the same baseline used in \cite{shenaj2023asynchronous}, where PASS~\cite{zhu2021prototype}, a continual learning strategy, is adapted to the federated scenario by combining it with FedAvg. In both cases, incorporating the rewind strategy results in enhanced performance, while maintaining computational costs low.

\section{Conclusion}

In this paper, we introduce \emph{FedRewind}, a novel approach that incorporates the \emph{rewind} technique into federated learning (FL) scenarios to address challenges arising from non-i.i.d. data distributions across distributed nodes. By periodically exchanging and rewinding models among nodes, \emph{FedRewind} mitigates issues related to overfitting on locally skewed data, which can hinder model generalizability and lead to catastrophic forgetting. This method significantly enhances performance by promoting robustness against class imbalances and improving overall model generalization, even in the complex context of Asynchronous Federated Continual Learning (AFCL).

We first validated \emph{FedRewind} on standard federated learning scenarios, demonstrating significant improvements in performance and generalization over existing methods such as FedAVG, CWT, and RWT. Importantly, these improvements were achieved without increasing computational costs, facilitating seamless integration into existing FL frameworks. We further evaluated our approach in the more extreme context of AFCL, surpassing existing methods. 

In conclusion, by integrating concepts from continual learning and leveraging  cognitive neuroscience principles, \emph{FedRewind} reduces the impact of distribution shifts, providing a robust solution to the challenges posed by non-i.i.d. data in distributed learning environments.

\section*{Acknowledgements} We acknowledge the support of the PNRR ICSC National Research Centre for High Performance Computing, Big Data and Quantum Computing (CN00000013), “Federated Learning as a Service (FLaaS) with Generative AI – GenF - Innovation Grant”, under the NRRP MUR program funded by the NextGenerationEU. Matteo Pennisi is a PhD student enrolled in the National PhD in Artificial Intelligence, XXXVII cycle, course on Health and life sciences, organized by Università Campus Bio-Medico di Roma.

\bibliographystyle{splncs04}
\bibliography{egbib}

\end{document}